\documentclass[12pt]{article}

\usepackage{scicite}
\PassOptionsToPackage{hyphens}{url}\usepackage{hyperref}%
\usepackage{graphicx}
\usepackage{xcolor}
\usepackage{times}
\usepackage{amsmath,amssymb}

\usepackage{soulpos}

\usepackage{tikz}
\usetikzlibrary{matrix,chains,arrows.meta,bending,positioning,quotes,shapes.geometric}
\usepackage{lineno}

\title{Passed the Turing Test: Living in Turing Futures}

\author
{Bernardo Gonçalves$^{1,2,\href{https://orcid.org/0000-0003-2794-8478}{\includegraphics[scale=0.08]{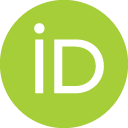}}}$\\
\\
\normalsize{$^{1}$Center for Artificial Intelligence (C4AI) and Polytechnic School,}\\ \normalsize{University of São Paulo, Brazil}\\
\normalsize{$^{2}$King's College, University of Cambridge, UK}
}

\date{}

\begin{document} 

\baselineskip24pt


\maketitle

\noindent
The world has seen the emergence of machines based on pretrained models, transformers, also known as generative artificial intelligences for their ability to produce various types of content, including text, images, audio, and synthetic data. Without resorting to preprogramming or special tricks, their intelligence grows as they learn from experience, and to ordinary people, they can appear human-like in conversation. This means that they can pass the Turing test, and that we are now living in one of many possible Turing futures where machines can pass for what they are not. However, the learning machines that Turing imagined would pass his imitation tests were machines inspired by the natural development of the low-energy human cortex. They would be raised like human children and naturally learn the ability to deceive an observer. These ``child machines,'' Turing hoped, would be powerful enough to have an impact on society and nature.

\section*{What Is the Turing Test?}
In 1950 \cite{turing1950}, Alan Turing proposed to replace the question `Can machines think?' with a new question based on what he called ``the imitation game'' (p.~433) and his ``test'' (pp.~446-447, 454). In its most familiar form, the new question was whether a machine playing A, the deceiver, could imitate B, the human assistant, in a remotely played conversation game to pass as B in the eyes of an average human interrogator playing C, the judge.

Note that the challenge of conscious impersonation and deception on the machine side slips into the problem of mere indistinguishability in the judgment of the human playing C, ``who should not be expert about machines'' \cite{turing1952a}.  
It is often asked why Turing made player C a nonspecialist, thus making the test somewhat easier for the machine. Looking at the historical context \cite{goncalves2024beautiful,goncalves2023argument}, we see that he was challenging his critics' confidence in the unassailable superiority of humans over other species in nature, to which he half-seriously included machines. 
For example, Turing referred to intelligent machines in a letter as ``another species of the thinking genus'' \cite{goncalves2024beautiful}. 
Having an ordinary human identify the machine in a blind test seems like a lighthearted way of posing that challenge. To illustrate that, Turing had his imaginary machine mimic human cultural stereotypes. For example, it would pose as an enlightened English individual who could compose a Shakespearean sonnet and discuss it metaphorically in relation to Mr. Pickwick \cite[p.~446]{turing1950}, a character in the literary work of the English writer Charles Dickens.

\section*{The Turing Test and Early AI}
Turing's test inspired early artificial intelligence (AI) scientists and has long been the most widely recognized criterion for machine intelligence. 
John McCarthy and Claude Shannon referred to it in their collection \emph{Automata Studies} (Princeton, 1956),  as ``the Turing definition of thinking'' and Turing's ``strong criterion.''
Together with Marvin Minsky and Nathaniel Rochester, McCarthy and Shannon in 1955 defined ``the AI problem'' as ``that of making a machine behave in ways that would be called intelligent if a human were so behaving'' \cite[p.~7]{mccarthy1955}. 
In 2013, when asked about Turing's test in a taped interview, Minsky said: ``The Turing test is a joke, sort of, about saying `A machine would be intelligent if it does things that an observer would say must be being done by a human.'$\,$''$\,$ 
This materially connects the early definition of ``the AI problem'' to Turing's test.

In the late 1960s, Minsky advised Stanley Kubrick and Arthur Clarke on their screenplay \emph{2001: A Space Odyssey}, which featured the Turing test-passing HAL: 

\begin{quotation}\noindent
``The sixth member of the crew cared for none of these things, for it was not human. It was the highly advanced HAL 9000 computer, the brain and nervous system of the ship $\hdots$ 
Whether HAL could actually think was a question which had been settled by the British mathematician Alan Turing back in the 1940s $\hdots$ Turing had pointed out that, if one could carry out a prolonged conversation with a machine $\hdots$ without being able to distinguish between its replies and those that a man might give, then the machine was thinking, by any sensible definition of the word $\hdots$ HAL could pass the Turing test with ease.''
\end{quotation}

\noindent
After HAL, Turing's test would become legendary.

Every time AI succeeds in automating a new task that would require intelligence if performed by humans, ``the Turing definition'' conquers new territory, and the importance of Turing's early message becomes clearer. 
The elegance of the Turing test definition, and the reason it has stood the test of time, lies in Turing's observation that human intelligence itself was largely unknown, and would likely remain so for some time. 
He was responding to one of his contemporaries who quoted René Descartes to argue for the special place of the human brain and language in nature \cite[Ch.~4]{goncalves2023argument}.  
Especially in the absence of a widely accepted definition of human intelligence, it is not surprising that machine intelligence will ultimately be judged by the tasks it can perform.

\section*{A Thought Experiment}
\begin{figure}[t]
\centering
\includegraphics[width=.80\textwidth]{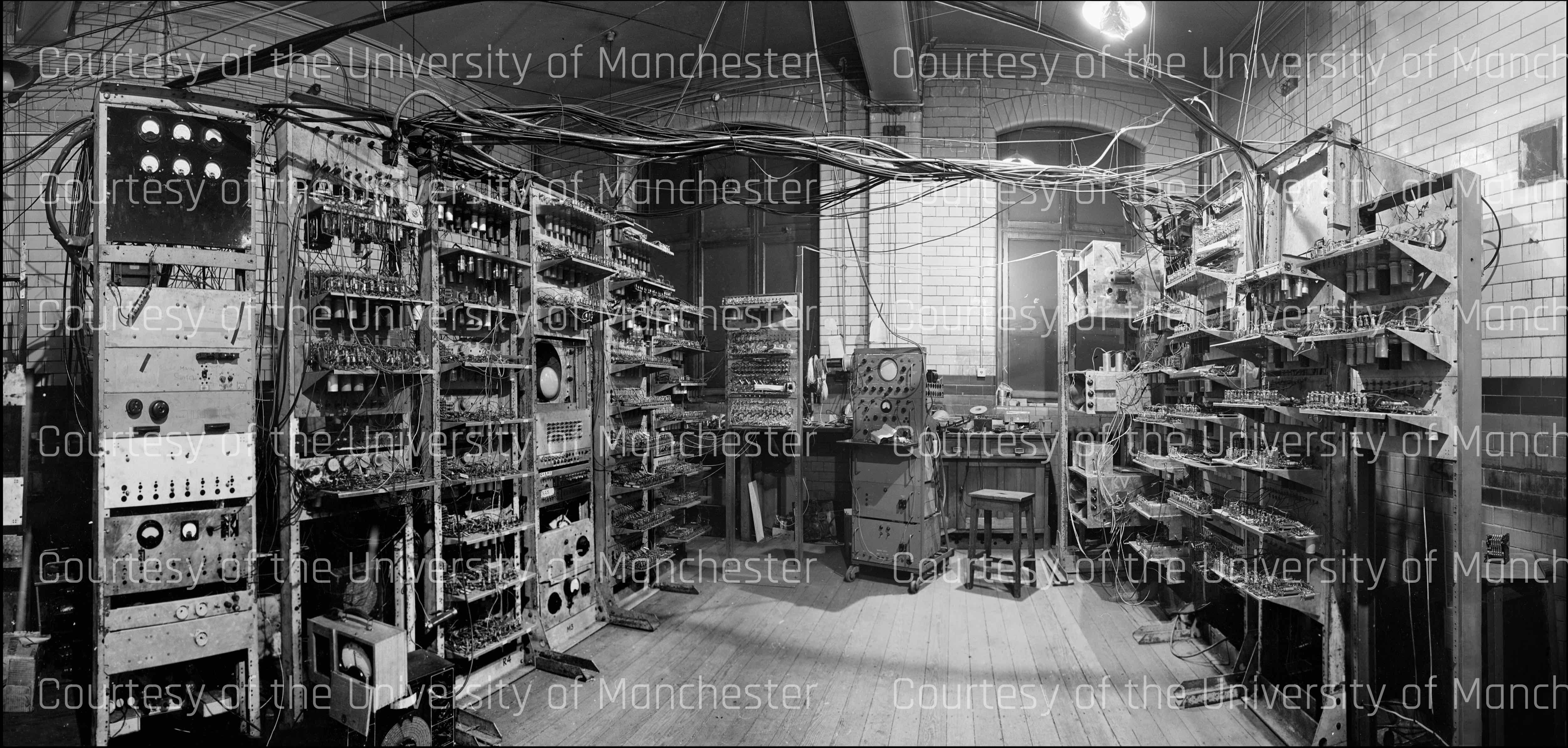}
\caption{University of Manchester Mark I, December 1948. A digital computer whose primary memory was on the order of $10^3$ units and whose programming system was designed by Turing. Courtesy of The University of Manchester.} 
\label{fig:mark1}
\end{figure}

Turing predicted that ``at the end of the century'' a learning machine would be able to play the imitation game well and pass the test, and that talk of ``machines thinking'' would be commonplace in ``the general educated opinion.'' 
At the time he made these two predictions \cite[p.~442]{turing1950}, a computer storage capacity of $10^9$ units was still fanciful speculation (see Fig.~\ref{fig:mark1}).   

However, instead of specifying a controlled experiment, Turing continuously varied the conditions of his test (having player B as a woman, a man, another machine, etc.), and in effect used it as a thought experiment to argue for machine intelligence \cite[Ch.~5]{goncalves2023argument}.
He also wrote that the ``only really satisfactory support'' that can be given for the two predictions would be ``doing the experiment described'' \cite[p.~455]{turing1950}. Did he mean recruiting women, men, machines, etc. for a practical ``imitation game''? No, absolutely not. The gender and the machine-versus-human elements were a half-serious way of responding to his critics \cite{goncalves2024beautiful}. Turing immediately shifted the focus to research on learning machines: ``What steps should be taken now if the experiment is to be successful?'' \cite[p.~455]{turing1950}. The rhetoric of his end-of-the-century experiment can be best understood as part of his propaganda for what he thought, writing in 1948 \cite{turing1948}, ``would probably have some effect'' in convincing critics and opponents: ``the actual production of machines.'' This would be the realization of his thought experiment.  

If anyone still wanted to seriously pursue a literal reading of Turing's imitation game and perform practical ``Turing'' tests using human participants, the research of Joseph Weizenbaum in the mid-1960s suggested that there was no point in doing so \cite{weizenbaum1966}. Using preprogramming and psychological methods, Weizenbaum presented evidence that people's attitudes towards talking machines may depend not only on the machines' behavior, but also on their own drives and prior assumptions. Turing had actually noted this in 1948 \cite{turing1948}: ``The extent to which we regard something as behaving in an intelligent manner is determined as much by our own state of mind and training as by the properties of the object under consideration.''

Like Galileo's falling-bodies experiment \cite[Ch.~6]{goncalves2023argument}, Turing's test was not presented as a practical experiment. However, this does not mean that Turing's ``test'' is a misnomer, for it has served as an empirical definition, criterion, and goal for machine intelligence ever since the early AI period\cite{goncalves2024beautiful}. 
In addition, as the physicist and historian Ernst Mach pointed out, ``thought experiment often precedes and prepares physical experiments'' \cite[Ch.~5]{goncalves2023argument}. 
This observation liberates scientists from Turing's original rhetoric, which belongs to his time and place. Instead, we can design modern Turing-like tests to evaluate the abilities of intelligent machines.

Before we explore this, we must return to the question: Can transformers pass the Turing test? That is, do they realize Turing's thought experiment? 
The short answer, as developed in the next two sections, is yes.

\section*{``The Nature of an Adequate Proof''} 

Weizenbaum wrote that his experiment was a ``striking form of Turing's test'' \cite[p.~42]{weizenbaum1966}. However, Turing was not interested in cheap deception and psychological tricks. He implied on several occasions that the ability of a machine to learn by itself was the key to an adequate proof of concept for machine intelligence, and thus the proper approach to preparing it for his test.   

The machines he envisioned in 1946 would be able to change their structure autonomously by learning from experience, like brains,  ``... by changing its neuron circuits'' through ``the growth of axons and dendrites'' \cite{turing1946}.  
Writing in 1950, he suggested that the ability to represent human-like fallibility should be acquired as a by-product of the learning process: 
``Another important result of preparing our machine for its part in the imitation game by a process of teaching and learning is that `human fallibility' is likely to be omitted [from the teaching] in a rather natural way, i.e., [learned] without special `coaching'$\,$'' $\hdots$ Processes that are learnt do not produce a hundred per cent, certainty of result; if they did they could not be unlearnt'' \cite[p.~459]{turing1950}. Overall, Turing did not conceive of machine intelligence without a learning foundation.

Later, in 1951, he further developed the link between learning and proving the concept of his test \cite{turing1951a}: 

\begin{quotation}\noindent 
My contention is that machines can be constructed which will simulate the behaviour of the human mind very closely $\hdots$
It would be the actual reaction of the machine to circumstances that would prove my contention, if indeed it can be proved at all $\hdots$ 
Let us go rather more carefully into the nature of this `proof.'  
\end{quotation}

\noindent
He continued: 

\begin{quotation}\noindent
It is clearly possible to produce a machine which would give a very good account of itself for any range of tests, if the machine were made sufficiently elaborate. However, this again would hardly be considered an adequate proof. Such a machine would give itself away by making the same sort of mistake over and over again, and being quite unable to correct itself, or to be corrected by argument from outside. If the machine were able in some way to `learn by experience' it would be much more impressive. If this were the case there seems to be no real reason why one should not start from a comparatively simple machine, and, by subjecting it to a suitable range of `experience' transform it into one which was more elaborate, and was able to deal with a far greater range of contingencies.   
\end{quotation}

\noindent
Turing then distinguished the above method based on learning from preprogramming:   

\begin{quotation}\noindent 
This process could probably be hastened by a suitable selection of the experiences to which it was subjected. This might be called `education.' But here we have to be careful. It would be quite easy to arrange the experiences in such a way that they automatically caused the structure of the machine to build up into a previously intended form, and this would obviously be a gross form of cheating, almost on a par with having a man inside the machine.   
\end{quotation}

\noindent
This explains why Turing saw research on ``learning machines'' as the steps that ``should be taken if the experiment [his test] is to be successful'' \cite[p.~455]{turing1950}. Preprogrammed machines are irrelevant to the test, as they are the same as having ``a man inside.'' They are prone to ``Lady Lovelace's objection'' \cite{goncalves2024lovelace}, ``which stated that the machine can only do what we tell it to do.'' For this reason, preprogrammed machines could never be a solid technological basis for making talk of ``machine thinking'' commonplace in the ``general educated opinion'' \cite[p.~442]{turing1950}.

\section*{Turing Test Passed}

The ``attention'' mechanism \cite{vaswani2017,perez2021} of the transformer architecture, and its AI systems, has yielded important empirical results in the imitation of human behavior. Hype aside, it has laid the groundwork for proving the concept associated with Turing's test argument: the existence of machines built with a relatively simple logical structure, whose intelligence grows by learning from experience to perform well at tasks once thought to be the province of humans in nature. Moreover, they can do so without resorting to preprogramming and Weizenbaum-style tricks.   

Importantly, the intelligence of generative transformers grows with the scaling of the model and its pretraining data \cite{brown2020}.  
There are tasks where their ability can be seen to increase gradually, and tasks where their ability emerges at a critical scale \cite{bigbench2023}, typically because the latter involve brittle metrics. This happens, for example, in arithmetic tasks such as addition, which Turing used to illustrate that machines should be allowed to make mistakes, especially as they learn. Presenting the imitation game, he made the machine miscalculate the addition of 34,957 to 70,764: ``(Pause about 30 seconds and then give as answer) 105621'' \cite[p.~434]{turing1950}. 
Transformers can eventually learn to add numbers without being taught directly. However, because an addition is either right or wrong, the ability suddenly appears at a critical scale. 

This is a phenomenon that Turing himself postulated for machine intelligence. In discussing how machine learning addresses ``Lady Lovelace's objection,'' he wrote \cite[p.~454]{turing1950}:  

\begin{quotation}
\noindent
One could say that a man can `inject' an idea into the machine, and that it will respond to a certain extent and then drop into quiescence, like a piano string struck by a hammer $\hdots$ Another simile would be an atomic pile of less than critical size: an injected idea is to correspond to a neutron entering the pile from without $\hdots$ Each such neutron will cause a certain disturbance which eventually dies away $\hdots$ If, however, the size of the pile is sufficiently increased, the disturbance caused by such an incoming neutron will very likely go on and on increasing until the whole pile is destroyed $\hdots$ Is there a corresponding phenomenon for minds, and is there one for machines? There does seem to be one for the human mind.  
\end{quotation}

\noindent
Although the intelligence of transformers increases as they gain experience and may eventually reach a critical point to acquire a skill, this does not mean that they are capable of conscious impersonation and deception. It just means that they prove the concept of machine intelligence in a way that is close enough to what Turing considered an ``adequate proof'' in connection with his test. 

It is often said that transformers can only remember, and imitation does indeed require memorization. However, sustaining an extended conversation requires some intelligence beyond mere memorization. To see this, note that a purely memorizing machine would have to store answers to every possible scenario presented by an interlocutor, including their combinations, as in Turing's example of associating a sonnet with a literary character. This would require storage exponential in the length of the test. Following a similar observation, and considering the Planck constant, Stuart Shieber used an estimate of the information capacity of the entire known universe as an upper bound on memory. Based on memory alone, even such an extremely large machine could only sustain a conversation for less than a minute \cite{shieber2014}.

In any case, the learning efficiency of transformers is still very low, limited by a low-level architecture. On learning efficiency, Turing wrote \cite[p.~457]{turing1950}: 

\begin{quotation}
\noindent 
The use of punishments and rewards can at best be a part of the teaching process. Roughly speaking, if the teacher has no other means of communicating to the pupil, the amount of information which can reach him does not exceed the total number of rewards and punishments applied. By the time a child has learnt to repeat `Casabianca' he would probably feel very sore indeed, if the text could only be discovered by a `Twenty Questions' technique, every `NO' taking the form of a blow. It is necessary therefore to have some other `unemotional' channels of communication. If these are available it is possible to teach a machine by punishments and rewards to obey orders given in some language, e.g., a symbolic language.  
\end{quotation}

\noindent
That is, although Turing saw learning as the basis of all intelligent computing, he also considered the efficiency and control of the learning process. In analogy to the human child, machine learning should scale and reach higher levels of abstraction through the use of language.  

Here, sensorimotor technologies and multimodal approaches using multiple data modalities and embedding mechanisms open new possibilities, as Turing hoped, ``to provide the machine with the best sense organs that money can buy,'' and allow the learning process to ``follow the normal teaching of a child ... Things would be pointed out and named, etc.'' \cite[p.~460]{turing1950}.

\section*{Living in Turing Futures}

For Turing, narrowing the gap ``between what machine and brain can do'' was indeed, as he wrote in a 1951 letter \cite{goncalves2024beautiful}, ``largely a quantitative matter.'' However, hardware capacity alone, he noted, would not be enough: ``Perhaps we may have enough capacity, but just won’t find an appropriate programme.''  
Writing about computer architecture in 1946, Turing expressed suspicion of the ``tradition of solving one's difficulties by means of much equipment rather than by thought'' \cite[p.~352]{hodges1983}. 
Several years earlier in 1925, he had written: 
``I always seem to want to make things from the thing that is commonest in nature \& with the least waste in energy'' \cite[p.~19]{hodges1983}.  
Turing followed such naturalistic principles. He sought the natural and social development of the relatively low-energy cortex of a human child as a target model for learning machines \cite[p.~457]{turing1950}. In contrast, generative AI overexploits natural resources by consuming unsustainable amounts of computing power. As it becomes a general-purpose technology, AI must become sustainable by moving from power-hungry machine learning to Turing's nature-inspired science.

Further, when early digital computers were on the verge of replacing human computers, who were mostly women, the computer pioneers, with one exception, were oblivious to this near-term prospect. The exception was Alan Turing.    
After Douglas Hartree tried to reassure the public that computers would not automate ``thought,'' only ``labor,'' Turing responded that ``the masters,'' not just ``the servants,'' were also ``liable to get replaced'' \cite{goncalves2024lovelace}. He feared that those in positions of power would try to undermine intelligent machines in order to maintain their dominance. He suggested that automation should affect people equally in society, not continually displace the lower class of workers and benefit only a few owners of the means of production. Against the social division of labor, his position was not far from the idea that the wealth created by intelligent machines should be nationalized or socialized \cite{xiang2018}.

Turing's suspicions about the uses of machines under the control of a few may have helped to motivate his conception of ``child machines'' \cite{turing1950}. They would be able to impersonate different profiles to deceive people when necessary, such as when they are supposed to prove their intelligence. His vision is in part a warning against our anthropocentrism and over-dominant place in nature. Pushed to its limits, it may find a representation in \emph{2001}'s HAL.

Whether to prevent dystopian futures or to steer towards utopian ones, the question of AI evaluation seems critical, and here again we can look to Turing for inspiration.

\section*{Turing-like AI Testing} 

\begin{figure}[t]
\centering
\includegraphics[width=.80\textwidth]{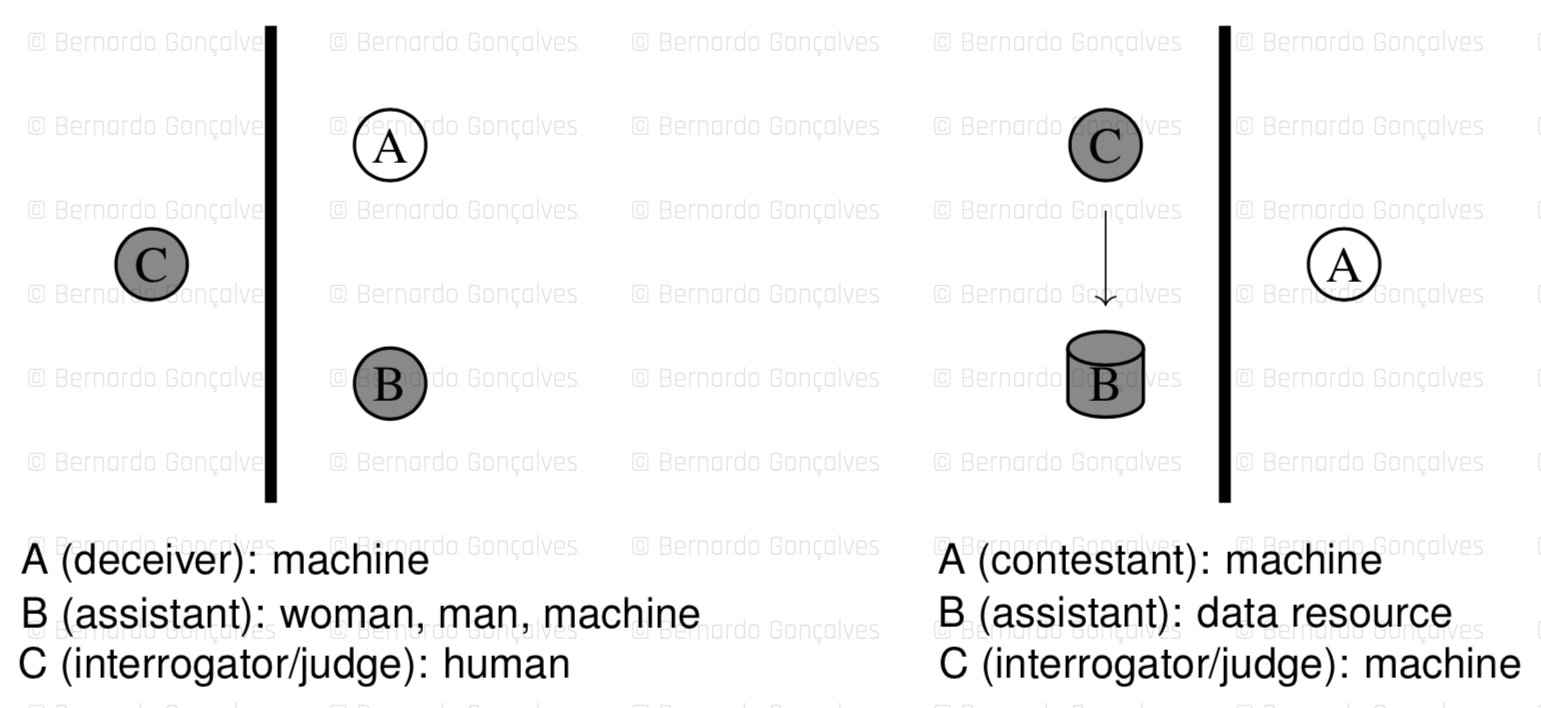}

\caption{Turing's original test (left): C is an ordinary human working with the help of another human, B, to correctly identify A, a machine that is trying to imitate and pass itself off as B in the eyes of C. 
Modern Turing-like test for AI evaluation (right): C is a machine that rigorously evaluates the abilities of A, an AI, supported by a data resource B. In both scenarios, the gray colored players play against the white colored machine. }
\label{fig:test}
\end{figure}

As AI systems are increasingly deployed in high-stakes scenarios, we may need to move beyond aggregate metrics and static benchmarks of input-output pairs, such as the Beyond the Imitation Game Benchmark (BIG-bench) \cite{bigbench2023}. We should be prepared to evaluate an AI's cognitive abilities in a way that resembles the realistic settings in which it will be used. This can be done with modern Turing-like tests (see Fig.~\ref{fig:test}).

A first element to consider for this research direction is the introduction of adversarial testing, as in Turing's test, but without any human players involved \cite{orallo2020}. A second element is the design of statistical protocols, as first proposed in the 1990s for Turing-like tests based on interactive proofs \cite{bradford1995}. These are probabilistic proofs designed to impose an asymmetry that exploits the computational resources of the prover (in our case, player A) relative to the verifier (player C), thus preventing the latter from being gamed. Turing-like AI testing could be a robust approach to emerging problems such as data contamination (the machine playing C should be able to use data retrieval and augmentation to generate challenging new instances at test time) and poisoning (the machine playing C should statistically cover the data domain at scale and thus be able to detect training vulnerabilities in the AI under test).    

Turing hoped ``that machines [would] eventually compete with men in all purely intellectual fields'' \cite[p.~460]{turing1950}. However, he left us with his test, now stripped of its discursive and rhetorical elements and considered at the level of conceptual foundations, as in the early days of AI.

\section*{Acknowledgments}
The author thanks Andrew Hodges for his comments on an earlier version of the manuscript, Fabio Cozman and Murray Shanahan for their support, the editor Sheng Jiang and the anonymous reviewers for their valuable comments, and Zhejiang Lab for the APC coverage. The author is solely responsible for the accuracy of this work.\\ 
\textbf{Funding:} The author thanks the Center for Artificial Intelligence (C4AI-USP), the São Paulo Research Foundation (FAPESP grants nos. 2019/07665-4, 2019/21489-4, and 2022/16793-9), and the IBM Corporation for their support. This article is a result of the project ``The Future of Artificial Intelligence: The Logical Structure of Alan Turing's Argument'' (2020-2024).\\ 
\textbf{Competing interests:} The author declares that he has no competing interests.

\bibliography{turing-argument-jul2024}
\bibliographystyle{Science}

\end{document}